\documentclass[10pt,twocolumn,letterpaper]{article}

\usepackage{iccv}
\usepackage{times}
\usepackage{epsfig}
\usepackage{graphicx}
\usepackage{amsmath}
\usepackage{amssymb}
\usepackage{subfigure}
\usepackage{scrextend}
\usepackage{multirow}
\usepackage[table]{xcolor}

\def\bb{{\bf b}}

\def\f{{\bf f}}

\def\W{{\bf W}}

\def\x{{\bf x}}
\def\y{{\bf y}}

\def\0{{\bf 0}}
\def\1{{\bf 1}}

\def\vecd{\mathrm{vec}}
\newcommand{\row}[2] {#1_{#2 \cdot}}
\newcommand{\col}[2] {#1_{\cdot #2}}

\usepackage[breaklinks=true,bookmarks=false]{hyperref}

\iccvfinalcopy % *** Uncomment this line for the final submission

\def\iccvPaperID{820} % *** Enter the ICCV Paper ID here

\ificcvfinal\fi
\begin{document}

%%%%%%%%% TITLE
\title{Factorized Bilinear Models for Image Recognition}

\author{Yanghao Li$^1$~~~~~ Naiyan Wang$^2$~~~~~ Jiaying Liu$^1$\thanks{Corresponding author}~~~~~ Xiaodi Hou$^2$\\
$^1$ Institute of Computer Science and Technology, Peking University~~~~~$^2$ TuSimple\\
{ \tt\small lyttonhao@pku.edu.cn~~ winsty@gmail.com~~ liujiaying@pku.edu.cn~~ xiaodi.hou@gmail.com}
% {\tt\small \{lyttonhao, liujiaying\}@pku.edu.cn}~~~{\tt\small \{winsty, shijianping5000, xiaodi.hou\}@gmail.com}
}
% \author{First Author\\
% Institution1\\
% Institution1 address\\
% {\tt\small firstauthor@i1.org}
% % For a paper whose authors are all at the same institution,
% % omit the following lines up until the closing ``}''.
% % Additional authors and addresses can be added with ``\and'',
% % just like the second author.
% % To save space, use either the email address or home page, not both
% \and
% Second Author\\
% Institution2\\
% First line of institution2 address\\
% {\tt\small secondauthor@i2.org}
% }

\maketitle
%\rowcolors{2}{white}{gray!25}
%\thispagestyle{empty}

\graphicspath{{figures/}}

%%%%%%%%% ABSTRACT
\begin{abstract}
Although Deep Convolutional Neural Networks (CNNs) have liberated their power in various computer vision tasks, the most important components of CNN, convolutional layers and fully connected layers, are still limited to linear transformations. In this paper, we propose a novel Factorized Bilinear (FB) layer to model the pairwise feature interactions by considering the quadratic terms in the transformations. Compared with existing methods that tried to incorporate complex non-linearity structures into CNNs, the factorized parameterization makes our FB layer only require a linear increase of parameters and affordable computational cost. To further reduce the risk of overfitting of the FB layer, a specific remedy called DropFactor is devised during the training process. We also analyze the connection between FB layer and some existing models, and show FB layer is a generalization to them. Finally, we validate the effectiveness of FB layer on several widely adopted datasets including CIFAR-10, CIFAR-100 and ImageNet, and demonstrate superior results compared with various state-of-the-art deep models. 
\end{abstract}

\begin{section}{Introduction}
Deep convolutional neural networks (CNNs)~\cite{alexnet,lecun1989backpropagation} have demonstrated their power in most computer vision tasks, from image classification~\cite{resnet,googlenet}, object detection~\cite{rcnn,faster-rcnn}, to semantic segmentation~\cite{fcn}. The impressive fitting power of a deep net mainly comes from its recursive feature transformations. Most efforts to enhance the representation power of a deep neural net can be roughly categorized into two lines. One line of works features on increasing the depth of the network, namely the number of non-linear transformations. ResNet~\cite{resnet} is a classic example of such extremely deep network. By using skip connections to overcome the gradients vanishing/exploding and degradation problems, ResNet achieves significant performance improvements. The other line of efforts aims at enhancing the fitting power for each layer. For example, Deep Neural Decision Forests~\cite{kontschieder2015deep} was proposed to integrate differentiable decision forests as the classifier. In ~\cite{lin2015bilinear}, the authors modeled pairwise feature interactions using explicit outer product at the final classification layer. The main drawbacks of these approaches are that they either bring in large additional parameters (for instance, \cite{lin2015bilinear} introduces 250M additional parameters for ImageNet classification) or have a slow convergence rate (\cite{kontschieder2015deep} requires 10x more epochs to converge than a typical GoogLeNet~\cite{googlenet}).

% NIN~\cite{lin2013network} replaces the linear convolutional filters with multilayer perceptrons (MLP), which is a general nonlinear functin approximator to enhance local modeling. However, the fully connected layeres in MLP are still composed of linear transformations. In~\cite{kontschieder2015deep}, decision trees are unified as the final predictors with the deep CNNs in a stochastic and differentiable way. Since the decision forest provides more powerful ability to handle high dimensional nonlinear data, this network achieves good performance on large image classication datasets, \eg Imagenet. Considering the pariwise interactions between features, bilinear pooling combine the final convolutional feature maps of two networks via the outer product at each localion of the feature maps, which has shown potential for fine-grained recognition~\cite{lin2015bilinear} and face verification~\cite{roychowdhury2015face}. A drawback of these approaches is that they bring along large additional parameters or complicate optimization for the training of a CNN.

In this paper, we propose the Factorized Bilinear (FB) model to enhance the capacity of CNN layers in a simple and effective way. At a glance, the FB model can be considered as a generalized approximation of the Bilinear Pooling~\cite{lin2015bilinear}, but with two modifications. First, our FB model generalizes the original Bilinear Pooling to all convolutional and fully connected layers. In this way, all computational layers in CNN could have larger capacity with pairwise interactions. However, under the original settings of Bilinear Pooling, such generalization will lead to explosion of parameters. To mitigate this problem, we constrain the rank of all quadratic matrices. This constraint significantly reduces the number of parameters and computational cost, making the complexity of FB layer \emph{linear} with respect to the original conv/fc layer. Furthermore, in order to cope with overfitting, we propose a regularization method called \emph{DropFactor} for the FB model. Analogous to Dropout~\cite{dropout}, DropFactor randomly sets some elements of the bilinear terms to zero during each iteration in the training phase, and uses all of them in the testing phase. 

To summarize, our contributions of this work are three-fold:
\begin{itemize}
\item  We present a novel Factorized Bilinear (FB) model to consider the pairwise feature interactions with linear complexity. We further demonstrate that the FB model can be easily incorporated into convolutional and fully connected layers. %It makes the FB layers be easily optimized in an end-to-end manner in CNN.
\item We propose a novel method \emph{DropFactor} for the FB layers to prevent overfitting by randomly dropping factors in the training phase.
\item We validate the effectiveness of our approach on several standard benchmarks. Our proposed method archives remarkable performance compared to state-of-the-art methods with affordable complexity.
\end{itemize}

\end{section}

\begin{section}{Related Work}

The Tao of tuning the layer-wise capacity of a DNN lies in the balance between model complexity and computation efficiency. The naive, linear approach of increasing layer capacity is either adding more nodes, or enlarging receptive fields. As discussed in~\cite{chatfield2014return}, these methods have beneficial effect up to a limit. From a different perspective, PReLU~\cite{he2015delving} and ELU~\cite{clevert2015fast} add flexibilities upon the activation function at a minimal cost, by providing a single learned parameter for each rectifier. Besides activation functions, many other works tried to use more complex, non-linear models to replace vector/matrix operations in each layer. For instance, Network In Network (NIN)~\cite{lin2013network} replaced the linear convolutional filters with multilayer perception (MLP), which is proven to be a general function approximator~\cite{hornik1989multilayer}. The MLP is essentially stacked of fully connected layers. Thus, NIN is equivalent of increasing the depth of the network. In~\cite{kontschieder2015deep}, random forest was unified as the final predictors with the DNNs in a stochastic and differentiable way. This back-propagation compatible version of random forest guides the lower layers to learn better representation in an end-to-end manner. However, the large computation overload makes this method inappropriate for practical applications.

Before the invention of deep learning, one of the most common tricks to increase model capacity is to apply kernels~\cite{shawe2004kernel}. Although the computational burden of some kernel methods can go prohibitively high, its simplest form -- bilinear kernel is certainly affordable. In fact, many of today's DNN has adopted bilinear kernel and have achieved remarkable performance in various tasks, such as fine-grained classification~\cite{lin2015bilinear,gao2015compact}, semantic segmentation~\cite{carreira2012semantic}, face identification~\cite{chowdhury2016one}, and person re-identification~\cite{chen2015similarity}. %Bilinear models were first introduced to separate style and content for images~\cite{tenenbaum2000separating}.

\begin{subsection}{Revisiting Bilinear Pooling} 
In~\cite{lin2015bilinear}, a method called Bilinear Pooling is introduced. In this model, the final output is obtained by a weighted pooling of a global descriptor, which comes from the outer product of the final convolutional layer with itself\footnote{Although the original Bilinear Pooling supports input vectors from two different networks, there is little difference performance-wise. For simplicity, we only consider the bilinear model using identical input vectors in this paper.}:

\begin{equation}
\begin{aligned}
	\mathbf{z} &= \sum_{i\in\mathbb{S}} \mathbf{x}_i\mathbf{x}_i^T,\\
\end{aligned}
\end{equation}
where $\{\x_i | \x_i \in \mathbb{R}^n, i \in \mathbb{S}\}$ is the input feature map, $\mathbb{S}$ is the set of spatial locations in the feature map, $n$ is the dimension of each feature vector, and $\mathbf{z} \in \mathbb{R}^{n \times n}$ is the global feature descriptor. Here we omit the signed square-root and $l_2$ normalization steps for simplicity. Then a fully connected layer is appended as the final classification layer:
\begin{equation}
\begin{aligned}
	\mathbf{y} &= \mathbf{b} + \mathbf{W}\vecd(\mathbf{z}),
\end{aligned}
\end{equation}
where $\vecd(\cdot)$ is the vectorization operator which converts a matrix to a vector, $\mathbf{W} \in \mathbb{R}^{c \times n^2}$ and $\mathbf{b} \in \mathbb{R}^{c}$ are the weight and bias of the fully connected layer, respectively. $\mathbf{y} \in \mathbb{R}^{c}$ is the output raw classification scores, and $c$ is the number of classification labels.

It is easy to see that the size of the global descriptor can go huge. To reduce the dimensionality of this quadratic term in bilinear pooling, \cite{gao2015compact} proposed two approximations to obtain compact bilinear representations. Despite the efforts to reduce dimensionality in~\cite{gao2015compact}, bilinear pooling still has large amounts of parameters and heavy computation burden. In addition, all of these models are based on the interactions of the final convolution layer, which is not able to be extended to earlier feature nodes in DNN.
\end{subsection}

\end{section}

\begin{section}{The Model}
\rowcolors{2}{white}{gray!25}
Before introducing the FB models, we first rewrite the bilinear pooling with its fully connected layer as below:
\begin{equation}\label{eq:bilinear}
\begin{aligned}
y_j &= b_j + \row{\mathbf{W}}{j}^T\vecd(\sum_{i \in \mathbb{S}} \mathbf{x}_i \mathbf{x}_i^T)\\
	&= b_j + \sum_{i \in \mathbb{S}}\mathbf{x}_i^T\row{\mathbf{W}}{j}^{R}\mathbf{x}_i,
\end{aligned}
\end{equation}
where $\row{\mathbf{W}}{j}$ is the $j$-th row of $\W$, $\row{\mathbf{W}}{j}^{R} \in \mathbb{R}^{n \times n}$ is a matrix reshaped from  $\row{\mathbf{W}}{j}$, and $y_j$ and $b_j$ are the $j$-th value of $\y$ and $\bb$. Although the bilinear pooling is capable of capturing pairwise interactions, it also introduces a quadratic number of parameters in weight matrices $\row{\mathbf{W}}{j}^{R}$, leading to huge computational cost and the risk of overfitting.

Previous literatures, such as~\cite{zeiler2014visualizing} have observed patterns of the co-activation of intra-layer nodes. The responses of convolutional kernels often form clusters that have semantic meanings. This observation motivates us to regularize $\row{\mathbf{W}}{j}^{R}$ by its rank to simplify computations and fight against overfitting.

\begin{subsection}{Factorized Bilinear Model}\label{sec:fblayer}
Given the input feature vector $\mathbf{x} \in \mathbb{R}^{n}$ of one sample, a common linear transformation can be represented as:
\begin{equation}\label{eq:basic}
\begin{aligned}
y &= b +\mathbf{w}^T\mathbf{x},
\end{aligned}
\end{equation}
where $y$ is the output of one neuron, $b$ is the bias term, $\mathbf{w} \in \mathbb{R}^{n}$ is the corresponding transformation weight and $n$ is the dimension of the input features.
%The above formulation covers both fully-connected layer and convolutional layers. Specifically, for one convolutional layer, suppose the input feature map is $\mathbf{X}$, the kernel weight is $\mathbf{W}$ and the output feature map is $\mathbf{Y}$. One output scalar $\mathbf{Y}_{k,i,j}$, which is in the k-th channel of $\mathbf{Y}$ with the location $(i,j)$, can be defined as:
% \begin{equation}
% \begin{aligned}
% \mathbf{Y}_{k,i,j} = P_k(\mathbf{W})Q_{i,j}(\mathbf{X}),
% \end{aligned}
% \end{equation}
% where $P_k$ is the operator extracting the k-th channel weight from $\mathbf{W}$, $Q_{i,j}$ is the operator extracting the 3D patch from $\mathbf{X}$ , which corresponds to the output of the location $(i,j)$. Note that $P_k$ and $Q_i,j$ would also flatten the obtained matrix to the vector form. Thus, by setting $y=\mathbf{Y}_{k,i,j}$, $\mathbf{W_1}=P_k(\mathbf{W})$ and $x=Q_{i,j}(\mathbf{X})$ , Eq.~\ref{eq:basic} can also been derived for convolutional layers. 

% Note that for the convolutional layer, $\mathbf{x}$ represents one patch of the input feature map with corresponding kernel size, which is flatten to a vector.

To incorporate the interactions term, we present the factorized bilinear model as follows:
\begin{equation}\label{eq:fb}
\begin{aligned}
y &= b + \mathbf{w}^T\mathbf{x} + \mathbf{x}^T\mathbf{F}^T\mathbf{F}\mathbf{x},
\end{aligned}
\end{equation}
where $\mathbf{F} \in \mathbb{R}^{k \times n}$ is the interaction weight with $k \in \mathbb{N}_0^+$ factors. To explain our model more clearly, the matrix expression of Eq.~(\ref{eq:fb}) can be expanded as:
\begin{equation}\label{eq:flatten}
\begin{aligned}
y = b + \sum_{i=1}^{n}w_ix_i + \sum_{i=1}^{n}\sum_{j=1}^{n}\langle \col{\f}{i}, \col{\f}{j} \rangle x_ix_j,
\end{aligned}
\end{equation}
where $x_i$ is the $i$-th variable of the input feature $\mathbf{x}$, $w_i$ is $i$-th value of the first-order weight and $\col{\f}{i}$ is the $i$-th column of $\mathbf{F}$. $\langle \col{\f}{i}, \col{\f}{j} \rangle$ is defined as the inner product of $\col{\f}{i}$ and $\col{\f}{j}$, which describes the interaction between the $i$-th and $j$-th variables of the input feature vector.

% The Eq.~\ref{eq:fb} shows that our factorization bilinear layer is a general transformation layer, which can be acted as both fully-connected layers and convolutional layers.

\begin{paragraph}{End-to-End Training.}
During the training, the parameters in FB model can be updated by back-propagating the gradients of the loss $l$. Let $\partial l/\partial y$ be the gradient of the loss function with respect to $y$, then by the chain rule we have:
\begin{equation}
\begin{aligned}
\frac{\partial l}{\partial \mathbf{x}} &= \frac{\partial l}{\partial y}\mathbf{w} + 2 \frac{\partial l}{\partial y} \mathbf{F}^T\mathbf{F}\mathbf{x}）,\\
\frac{\partial l}{\partial \mathbf{F}} &= 2 \frac{\partial l}{\partial y} \mathbf{F}\mathbf{x}\mathbf{x}^T,\\
 \frac{\partial l}{\partial \mathbf{w}} &= \frac{\partial l}{\partial y} \mathbf{x},\quad \frac{\partial l}{\partial b} = \frac{\partial l}{\partial y}.
\end{aligned}
\end{equation}
Thus, the FB model applied in DNNs can be easily trained along with other layers by existing optimizers, such as stochastic gradient descent.
\end{paragraph}

\begin{paragraph}{Extension to Convolutional Layers.} The aforementioned FB model can be applied in fully connected layers easily by considering all the output neurons. Besides, the above formulations and analyses can also be extended to the convolutional layers. Specifically, the patches of the input feature map in the convolutional layers can be rearranged into vectors using \texttt{im2col} trick~\cite{jia2014learning,ren2015vectorization}, and convolution operation is converted to dense matrix multiplication like in fully connected layers. Most popular deep learning frameworks utilize this reformulation to calculate the convolution operator, since dense matrix multiplication could maximize the utility of GPU. Thus, the convolutional layer could also benefit from the proposed FB model.

\end{paragraph}

\begin{paragraph}{Complexity Analysis.}
According to the definition of the interaction weight $\mathbf{F}$ in Eq.~(\ref{eq:fb}), the space complexity, which means the number of parameters for one neuron in the FB model, is $O(kn)$. Although the complexity of na\"{i}ve computation of Eq.~(\ref{eq:flatten}) is $O(kn^2)$, we can compute the factorization bilinear term efficiently by manipulating the order of matrix multiplication in Eq.~(\ref{eq:fb}). By computing $\mathbf{F}\mathbf{x}$ and $\mathbf{x}^T\mathbf{F}^T$ first, $\mathbf{x}^T\mathbf{F}^T\mathbf{F}\mathbf{x}$ can be computed in $O(kn)$. Thus, the total computation complexity of Eq.~(\ref{eq:fb}) is also $O(kn)$. As a result, the FB model has linear complexity in terms of both $k$ and $n$ for the computation and the number of parameters. We will show the actual runtime of the FB layers in our implementation in the experiments section.
\end{paragraph}
\end{subsection}

\begin{subsection}{DropFactor}\label{sec:dropfactor}

\begin{figure*}[t]
\center{
\subfigure[Expanding FB model structure]{\includegraphics[width=0.23\linewidth]{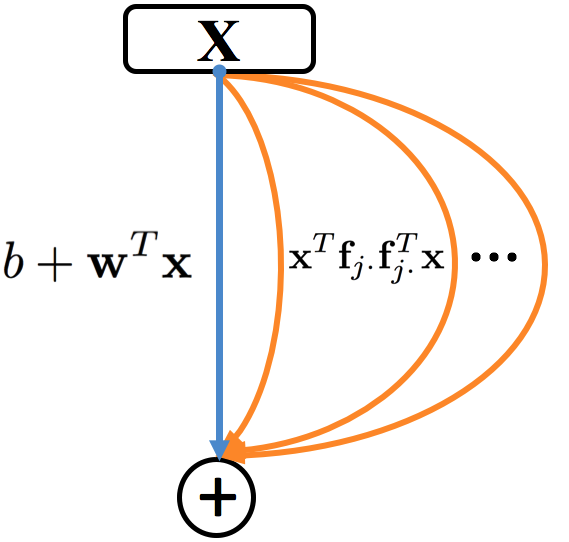}\label{fig:dropfactor-a} }~~~~~~~~
\subfigure[DropFactor at training time] {\includegraphics[width=0.23\linewidth]{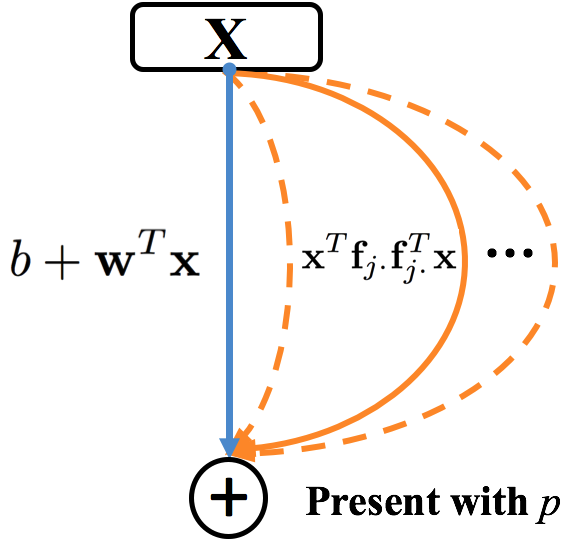}\label{fig:dropfactor-b}}~~~~~~~~
\subfigure[DropFactor at test time] {\includegraphics[width=0.23\linewidth]{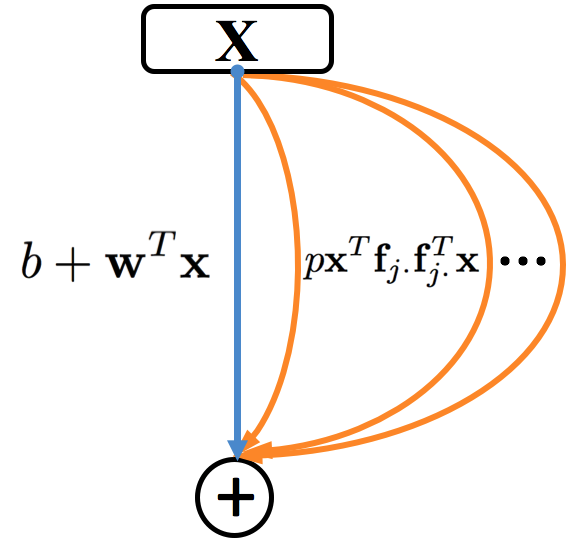}\label{fig:dropfactor-c}}
}
\caption{The structure of the FB layer and the explanations of DropFactor. (a) The expanding FB layer contains one conventional linear path (the blue line) and $k$ bilinear paths (the orange lines). (b) Each bilinear path is retained with probability $p$ at training time. (c) At testing time, each bilinear path is always present and the output is multiplied by $p$.}
\end{figure*}

Dropout~\cite{dropout} is a simple yet effective regularization to prevent DNNs from overfitting. The idea behind Dropout is that it provides an efficient way to combine exponentially different neural networks by randomly dropping neurons. Inspired by this technique, we propose a specific \emph{DropFactor} method in our FB model. 

We first reformulate Eq.~(\ref{eq:fb}) as:
\begin{equation}\label{eq:expand}
\begin{aligned}
y = b + \mathbf{w}^T\mathbf{x} + \sum_{j=1}^{k}\mathbf{x}^T\row{\mathbf{f}}{j}\row{\mathbf{f}}{j}^T\mathbf{x},\\
\end{aligned}
\end{equation}
where $\row{\mathbf{f}}{j}$ is the $j$-th row of interaction weight $\mathbf{F}$, which represents the $j$-th factor. Based on Eq.~(\ref{eq:expand}), Fig.~\ref{fig:dropfactor-a} shows the expanding structure of the FB layer which composes of one linear transformation path and $k$ bilinear paths. The key idea of our DropFactor is to randomly drop the bilinear paths corresponding to $k$ factors during the training. This prevents $k$ factors from co-adapting. 

In our implementation, each factor is retained with a fixed probability $p$ during training. With the DropFactor, the formulation of FB layer in the training becomes:
\begin{equation}\label{eq:drop}
\begin{aligned}
y = b + \mathbf{w}\mathbf{x} + \sum_{j=1}^{k}m_{j}\mathbf{x}^T\row{\mathbf{f}}{j}\row{\mathbf{f}}{j}^T\mathbf{x},
\end{aligned}
\end{equation}
where $m_j \sim \mathrm{Bernoulli}(p)$. With the DropFactor, the network can be seen as a set of $2^k$ thinned networks with shared weights. In each iteration, one thinned network is sampled randomly and trained by back-propagation as shown in Fig.~\ref{fig:dropfactor-b}.

% \begin{equation}
% \begin{aligned}
% y = w_0 + \sum_{i=1}^{n}w_ix_i + \sum_{i=1}^{n}\sum_{j=1}^{n}\langle \mathbf{m} * \mathbf{f_i}, \mathbf{m} \cdot \mathbf{f_j} \rangle x_ix_j,
% \end{aligned}
% \end{equation}
% where $*$ denotes element wise product and $\mathbf{m} \in \mathbb{R}^{1 \times k}$ is a binary mask in which each element $m_j \sim Bernoulli(p)$ is drawn with probability $p$.

For testing, instead of explicitly averaging the outputs from all $2^k$ thinned networks, we use the approximate ``Mean Network'' scheme in \cite{dropout}. As shown in Fig.~\ref{fig:dropfactor-c}, each factor term $\mathbf{x}^T\row{\mathbf{f}}{j}\row{\mathbf{f}}{j}^T\mathbf{x}$ is multiplied by $p$ at testing time:
\begin{equation}\label{eq:drop_test}
\begin{aligned}
y = b + \mathbf{w}^T\mathbf{x} + \sum_{j=1}^{k}p\mathbf{x}^T\row{\mathbf{f}}{j}\row{\mathbf{f}}{j}^T\mathbf{x}.
\end{aligned}
\end{equation}
In this way, the output of each neuron at testing time is the same as the expected output of $2^k$ different networks at training time.

% \begin{equation}
% \begin{aligned}
% y = w_0 + \mathbf{W}_1\mathbf{x} + \sum_{j=1}^{k}m_{j}\mathbf{x}^T\mathbf{c_j}^T\mathbf{c_j}\mathbf{x},
% \end{aligned}
% \end{equation}

% \begin{equation}
% \begin{aligned}
% y = w_0 + \mathbf{W}_1\mathbf{x} + \sum_{j=1}^{k}p\mathbf{x}^T\mathbf{c_j}^T\mathbf{c_j}\mathbf{x}
% \end{aligned}
% \end{equation}

\end{subsection}

\end{section}

\begin{section}{Relationship to Existing Methods}\label{sec:relation}
\rowcolors{2}{white}{gray!25}
In this section, we connect our proposed FB model with several closely related works, and discuss the differences and advantages over them.
\begin{figure}[t]
\center{
\subfigure[Bilinear Pooling block]{\includegraphics[width=0.4\linewidth]{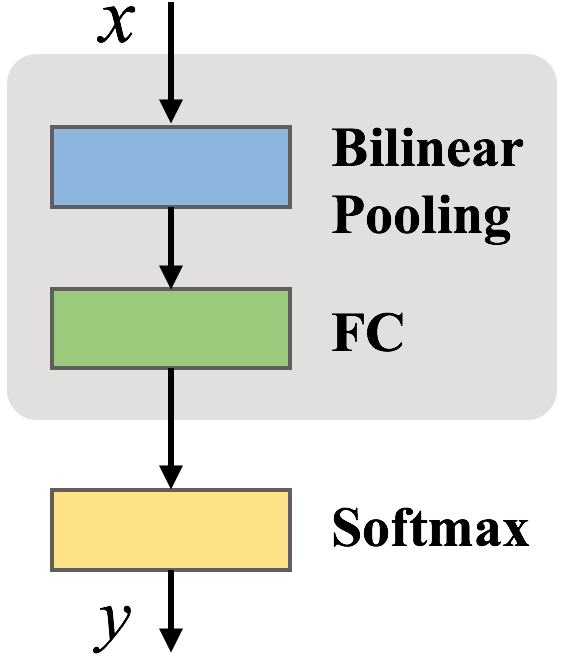}\label{fig:compare_bilinear-a} }~~~~~
\subfigure[Factorized Bilinear block] {\includegraphics[width=0.4\linewidth]{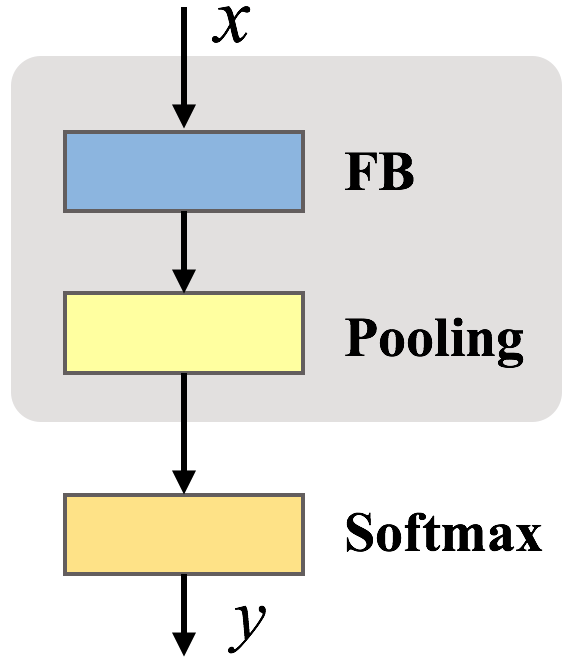}\label{fig:compare_bilinear-b}}
}
\caption{The structure of bilinear pooling block and its corresponding FB block. $\x$ is the input feature map of the final convolutional block.}
\end{figure}

\noindent\textbf{Relationship to Bilinear Pooling.}
Bilinear pooling~\cite{lin2015bilinear} modeled pairwise interactions of features by outer product of two vectors. In the following, we demonstrate that our FB block is a generalization form of bilinear pooling block.

% As shown in Fig~\ref{fig:compare_bilinear-a}, bilinear pooling is applied after the last convolutional layer of a CNN (\eg VGG) followed by the global pooling  layer and classification fully-connected layer. To be comparable with bilinear pooling, we can construct our Factorized Bilinear layer as the final prediction layer by using the 1x1 convolutional kernel as shown in Fig~\ref{fig:compare_bilinear-b}. Since the pooling layer can be considered as an adding operator and appeared in both bilinear pooling block and our Factorized Bilinear block, it can be ignored in the analysis for simplicity. Considering the feature vector $x$ in one location of the input conv feature map, the bilinear pooling block can be represented as:
% \begin{equation}
% \begin{aligned}
% y &= b + \mathbf{W_b}vec(\mathbf{x}\mathbf{x^T}) = b + \mathbf{x^T}\mathbf{W_b^{'}}\mathbf{x},
% \end{aligned}
% \end{equation}
% where $vec(\cdot)$ is the operator to scratch the matrix to a vector, $\mathbf{W_b} \in \mathbb{R}^{1*n^2}$ is the weight of the fully connected layer and $\mathbf{W_b^{'}} \in \mathbb{R}^{n*n}$ is transformed from $\mathbf{W_b}$. Thus, $\mathbf{W_b^{'}}$ is the weight of the interaction term like our $\mathbf{F^T}\mathbf{F}$ in Eq.~\ref{eq:fb}.

As shown in Fig.~\ref{fig:compare_bilinear-a}, the bilinear pooling is applied after the last convolutional layer of a CNN (\eg VGG) , then followed by a fully-connected layer for classification. We construct an equivalent structure with our FB model by using the FB convolutional layer with $1 \times 1$ kernel as shown in Fig.~\ref{fig:compare_bilinear-b}. The final average pooling layer is used to aggregate the scores around the spatial locations. Thus, Eq.~(\ref{eq:fb}) can be reformulated as:
\begin{equation}\label{eq:complete_fb}
\begin{aligned}
y &= b + \frac{1}{\| \mathbb{S} \|}\sum_{i \in \mathbb{S}}(\mathbf{w}^T\mathbf{x}_i + \mathbf{x}_i^T\mathbf{F}^T\mathbf{F}\mathbf{x}_i).
\end{aligned}
\end{equation}
Compared with bilinear pooling in Eq.~(\ref{eq:bilinear}), we add the linear term and replace the pairwise matrix $\row{\mathbf{W}}{j}^{R}$ with factorized bilinear weight $\mathbf{F}^T\mathbf{F}$.

%For any positive definite matrix $\mathbf{W}$, there exists a matrix $\mathbf{F}$ such that $\mathbf{W} = \mathbf{F^T}\mathbf{F}$ when $k$ is sufficiently large. Thus, the expressiveness of $\mathbf{F^T}\mathbf{F}$ is similar to that of $\row{\mathbf{W}}{j}^{'}$.
We argue that such symmetric and low rank constraints on the interaction matrix are reasonable in our case. First, the interaction between $i$-th and $j$-th feature and that between $j$-th and $i$-th feature should be same. Second, due to the redundancy in neural networks, the neurons usually form the clusters~\cite{zeiler2014visualizing}. As a result, only a few factors should be enough to capture the interactions between them. Besides reducing the space and time complexity, restricting $k$ also potentially prevents overfitting and leads to better generalization. %In addition, compared to the quadratic dimension of bilinear pooling features, our Factorized Bilinear layer significantly reduces the parameter memory and computation complexity.

An improvement of bilinear pooling is compact bilinear pooling~\cite{gao2015compact} which reduces the feature dimension of bilinear pooling using two approximation methods: Random Maclaurin (RM)~\cite{kar2012random} and Tensor Sketch (TS)~\cite{pham2013fast}. However, the dimension of the projected compact bilinear feature is still too large (10K for the 512-dimensional input) for deep networks. Table~\ref{tb:comp_complexity} compares the factorized bilinear with bilinear pooling and its variant compact bilinear pooling. Similar to compact bilinear pooling, our FB model requires much fewer parameters than bilinear pooling. It also reduces the computation complexity significantly (from 133M in TS to 10M) at the same time. In addition, not only used as the final prediction layer, our method can also be applied in the early layers as a common transformation layer, which is much more general than the bilinear pooling methods.

\begin{table}[t]
\small
	\begin{center}
	\setlength{\tabcolsep}{3pt}
	\begin{tabular}{lcc}
		\hline	
		Method					& Parameter		& Computation 		\\
		\hline
		Bilinear~\cite{lin2015bilinear} 		& $cn^2$ [262M]		& O($cn^2$) [262M]\\
		RM~\cite{gao2015compact} 		& $2nd+cd$ [20M]		& O($cnd$) [5G]\\
		TS~\cite{gao2015compact}			& $2n+cd$ [10M]			& O($c(n+dlogd$)) [133M]		\\
		Factorized Bilinear						& $ckn$ [10M]		& O($ckn$) [10M]		\\
		\hline
	\end{tabular}
	\end{center}
	\caption{The comparison of number of parameters and computation complexity among the proposed factorized bilinear, bilinear pooling and compact bilinear pooling. Parameters $n$, $c$, $d$, $k$ correspond to the dimension of input feature, the dimension of output (number of classes), the projected dimension of compact bilinear pooling and the number of factors in factorized bilinear. Numbers in brackets indicate typical values of each method for a common CNN on a 1000-class classification task, \ie, $n=512$, $c=1,000$, $d=10,000$, $k=20$. Note that we omit the width and height of the input feature map for simplicity.}\label{tb:comp_complexity}

\end{table}

%\end{paragraph}

\noindent\textbf{Relationship to Factorization Machines.} Factorization Machine (FM)~\cite{rendle2010factorization} is a popular predictor in machine learning and data mining, especially for very sparse data. Similar to our FB model, FM also captures the interactions of the input features in a factorized parametrization way. However, since FM is only a classifier, its applications are restricted in the simple regression and classification. In fact, a 2-way FM can be constructed by a tiny network composed of a single FB layer with one output unit. In this way, a 2-way FM is a special case of our FB model. While our FB model is much more general, which can be integrated into regular neural networks seamlessly for different kinds of tasks.

%\end{paragraph}

\end{section}

\begin{section}{Experiments}
\rowcolors{2}{white}{gray!25}
In this section, we conduct comprehensive experiments to validate the effectiveness of the proposed FB model.
In Sec.~\ref{exp:ablation}, we first investigate the design choices and properties of the proposed FB model, including the architecture of the network, parameters setting and speed. Then, we conduct several experiments on three well-known standard image classification datasets and two fine-grained classification datasets, in Sec.~\ref{exp:results}. In the following experiments, we refer the CNN equipped with our FB model as Factorized Bilinear Network (FBN). Code is available at \textcolor{red}{\url{https://github.com/lyttonhao/Factorized-Bilinear-Network}}.

\noindent\textbf{Implementation Details.}
We adopt two standard network structures: Inception-BN~\cite{bn} and ResNet~\cite{he2016identity} as our baselines. Our FBN improves upon these two structures. Some details are elaborated below. %To take advantage of the existing effective network structures and compare fairly, we construct our FBNs based on these two networks, respectively. In Sec.~\ref{exp:ablation}, we will introduce FBNs with different architectures and their corresponding performance.
For one specified network and its corresponding FBN, we use all the same experiment settings (\eg the training policy and data augmentation), except two special treatments for FBNs. (i) To prevent the quadratic term in FB layer explodes too large, we change the activation before every FB layer from `ReLU'~\cite{relu} to `Tanh', which restricts the output range of FB layers. We do not use the power normalization and $l_2$ normalization in~\cite{lin2015bilinear}. The reason is that: 1) square root is not numerically stable around zero. 2) we do not calculate the bilinear features explicitly. (ii) We use the slow start training scheme, which shrinks the initialized learning rate of the FB layer by a tenth and gradually increases the learning rate to the regular level in several epochs (\eg 3 epochs). This treatment learns a good initialization and is beneficial for converging of FBNs, which is similar to the warmup step in~\cite{resnet}.

\begin{subsection}{Ablation Analyses}\label{exp:ablation}
In this section we investigate the design of architecture of FBNs and the appropriate parameters, such as the number of factors $k$ and the DropFactor rate $p$~\footnote{More exploration experiments can be found in supplemental material.}. Most of the following experiments are conducted on a simplified version of Inception-BN network\footnote{\label{fn:simple-inception}\url{https://goo.gl/QwVS3Z}} on CIFAR-100. Some details about the experiment settings, such as training policies and data augmentations, will be explained in Sec.~\ref{exp:results}.
\begin{figure}[t]
\center{
\includegraphics[width=0.65\linewidth]{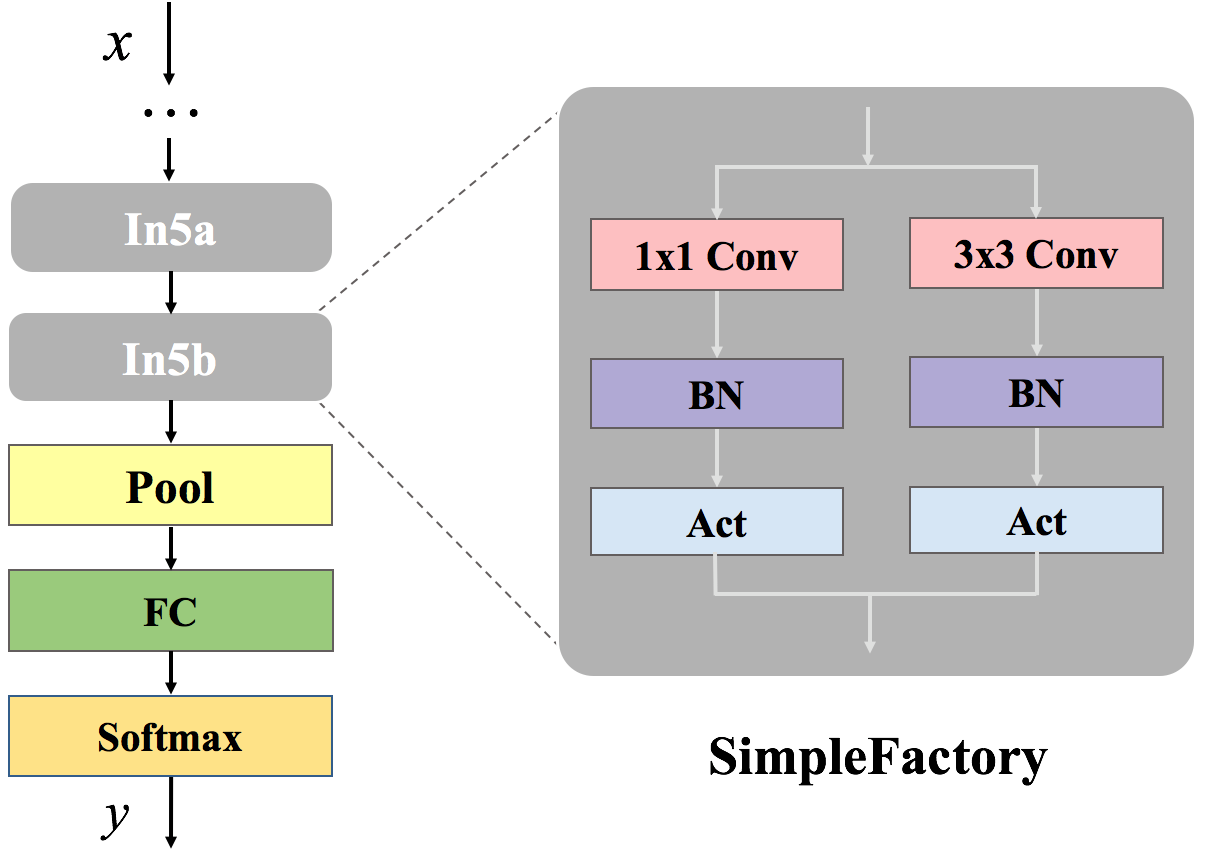}}
\caption{The structure of the simplified version of Inception-BN and the SimpleFactory block.\label{fig:inception}}
\vspace{-3mm}
\end{figure}
\begin{paragraph}{Architecture of FBNs.}
As discovered in~\cite{zeiler2014visualizing}, the lower layers of CNNs usually respond to simple low-level features. Thus, linear transformation is enough to abstract the concept within images. Consequently, we modify the higher layers of Inception-BN network to build our FBNs. As shown in Fig.~\ref{fig:inception}, the original Inception-BN is constructed by several SimpleFactories, and each SimpleFactory contains a $1 \times 1$ conv block and a $3 \times 3$ conv block. The five FBNs with different structures are explained as follows:
\begin{enumerate}
\itemsep-0.3em
\item \textbf{In5a-FBN.} We replace the 1x1 conv layer in In5a factory with our FB convolutional layer. The parameters such as kernel size and stride size are kept the same.
\item \textbf{In5b-FBN.} This is same as In5a-FBN except we apply FB model in In5b factory.
\item \textbf{FC-FBN.} The final fully-connected layer is replaced by our FB fully connected layer.
\item \textbf{Conv-FBN.} As shown in Fig.~\ref{fig:compare_bilinear-b}, Conv-FBN is constructed by inserting a FB conv layer with $1 \times 1$ kernel before the global pooling layer and removing the fully-connected layer.
\item \textbf{Conv+In5b-FBN.} This network combines Conv-FBN and In5b-FBN. 
\end{enumerate}

\begin{table}[htbp]
\small
	\begin{center}
	\begin{tabular}{lcc}
		\hline	
		Network type							& $p$ 	& CIFAR-100	\\
		\hline
		Inception-BN 				& -		&24.70\\
		In5a-FBN	& 0.8	& 24.73\\
		In5b-FBN 	& 0.8	& 22.63\\
		FC-FBN		& 0.5	& 24.07\\
		Conv-FBN	& 0.5	& \textbf{21.98}\\
		In5b+Conv-FBN &  (0.8, 0.5) 	& 23.70\\
		\hline
	\end{tabular}
	\end{center}	
	\caption{Test error (\%) of original Inception-BN and five FBNs on CIFAR-100. $p$ is the DropFactor rate. The $(0.8, 0.5)$ of $p$ in the last row indicates that $p$ is set as $0.8$ in In5b factory and $0.5$ in the Conv FB layer.} \label{tbl:fbn_net}
\vspace{-3mm}
\end{table}

\begin{figure}[htbp]
\center{
\includegraphics[width=0.8\linewidth]{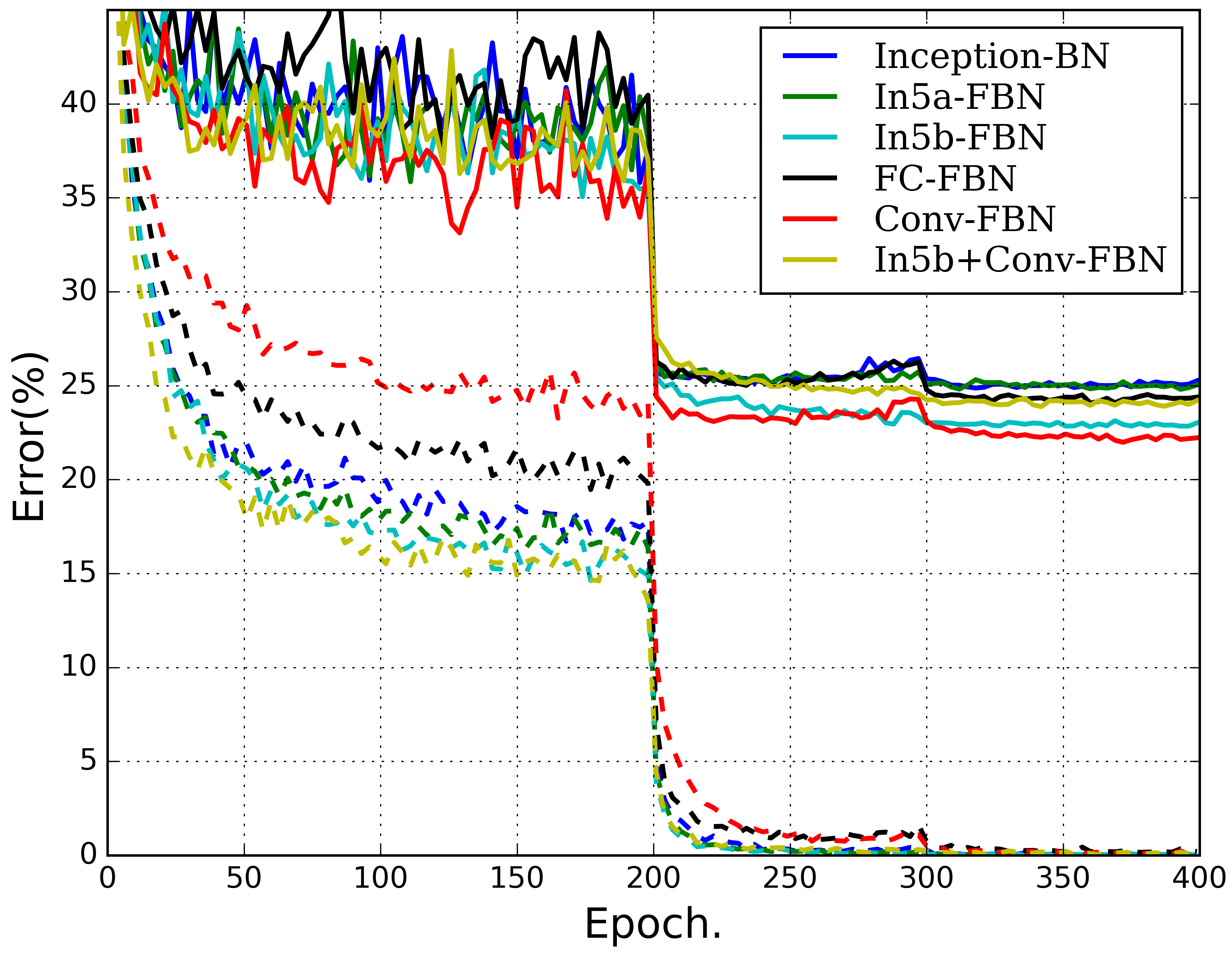}
}
\caption{Training on CIFAR-100 with different architectures. Dashed lines denote training error, and bold lines denote testing error. Best viewed in color.}\label{fig:fbn_net}
\vspace{-3mm}
\end{figure}

The results of original Inception-BN and five FBNs are shown in Table~\ref{tbl:fbn_net}. The training and testing curves for these networks are presented in Fig.~\ref{fig:fbn_net}. The number of factors $k$ of different FBNs is fixed as $20$ and the appropriate values for the DropFactor rate $p$ are chosen for different FBNs (More experiments about $k$ and $p$ are shown in Table~\ref{tbl:factors} and Fig.~\ref{fig:result-drop}). From Table~\ref{tbl:fbn_net}, we can see that most FBNs achieve better results than the baseline Inception-BN model, and Conv-FBN achieves 21.98\% error which outperforms Inception-BN by a large margin of 2.72\%. It demonstrates that incorporating FB model indeed improves the performance of the network.

From Table~\ref{tbl:fbn_net} and Fig.~\ref{fig:fbn_net}, we have several interesting findings: 1) Comparing the results of Conv-FBN, In5b-FBN and In5a-FBN, we find that incorporating FB model in the lower layers may lead to inferior results and suffer from overfitting more. 2) The difference between FC-FBN and Conv-FBN is whether to consider the interactions across different locations of the input feature map. The results show that the pairwise interactions should be captured at each position of the input separately. 3) Incorporating two FB blocks (In5b+Conv-FBN) does not further improve the performance at least in CIFAR-100, but leads to more severe overfitting instead.

\end{paragraph}

\begin{paragraph}{Number of Factors in FB layer.} As the number of parameters and computational complexity in the FB layer increase linearly in the number of factors $k$, we also evaluate the sensitivity of $k$ in a FB layer. Table~\ref{tbl:factors} shows the results of In5b-FBN and Conv-FBN on CIFAR-100. As can be noticed, after $k$ grows beyond 20, the increase of performance is marginal, and too large $k$ may be even detrimental. Thus, we choose 20 factors in all the subsequent experiments.

\begin{table}[htbp]
\small
	\begin{center}
	\begin{tabular}{ccc}
		\hline	
		factors $k$			& In5b-FBN & Conv-FBN	\\
		\hline
		10		  	& 23.07 & 22.99\\
		20			& 22.63 & 21.98\\
		50 			& 22.82 & 21.90\\
		80			& 23.07 & 21.88\\
		\hline
	\end{tabular}
	\end{center}	
	\caption{Test error (\%) on CIFAR-100 of In5b-FBN and Conv-FBN with different number of factors. The DropFactor rate $p$ is 0.8 and 0.5 for In5b-FBN and Conv-FBN according to the performance.} \label{tbl:factors}
\vspace{-3mm}
\end{table}

\end{paragraph}

\begin{paragraph}{DropFactor in FBNs.}

We also vary the DropFactor rate $p$ to see how it affects the performance. Fig.~\ref{fig:result-drop-a} shows the testing error on CIFAR-100 of In5b-FBN and Conv-FBN with different $p$. Note that even the FBNs without DropFactor ($p = 1.0$) can achieve better results than the baseline method. With the DropFactor, FBNs further improve the result and achieve the best result $21.98\%$ when $p = 0.5$ for Conv-FBN and $22.63\%$ when $p = 0.8$ for In5b-FBN. Fig.~\ref{fig:result-drop-b} and \ref{fig:result-drop-c} show the training and testing curves with different $p$. As illustrated, the testing curves are similar at the first 200 epochs for different networks, yet the training curves differ much. The smaller DropFactor rate $p$ makes the network less prone to overfitting. It demonstrates the effectiveness of DropFactor. On the other hand, a too small rate may deteriorate the convergence of the FBNs.

% \begin{figure}[hbtp]
% \center{
% \includegraphics[width=0.9\linewidth]{dropfactor}
% \caption{Test error (\%) on CIFAR-100 of In5b-FB and Conv-FB networks with $k = 20$ and different drop rates.}\label{fig:dropfactor}
% }
% \end{figure}

% \begin{figure}[hbtp]
% \center{
% \includegraphics[width=0.9\linewidth]{curve_dropfactor}
% \caption{Training on CIFAR-100 of Conv-FB networks with $k = 20$ and different drop rates. Dashed lines denote training error, and bold lines denote testing error.We do not show the results of larger drop rates, since the performance drop significantly when $p$ is too large.}\label{fig:dropfactor}
% }
% \end{figure}

\begin{figure*}[t]
\center{
\subfigure[Test error of In5b-FBN and Conv-FBN]{\includegraphics[width=0.3\linewidth]{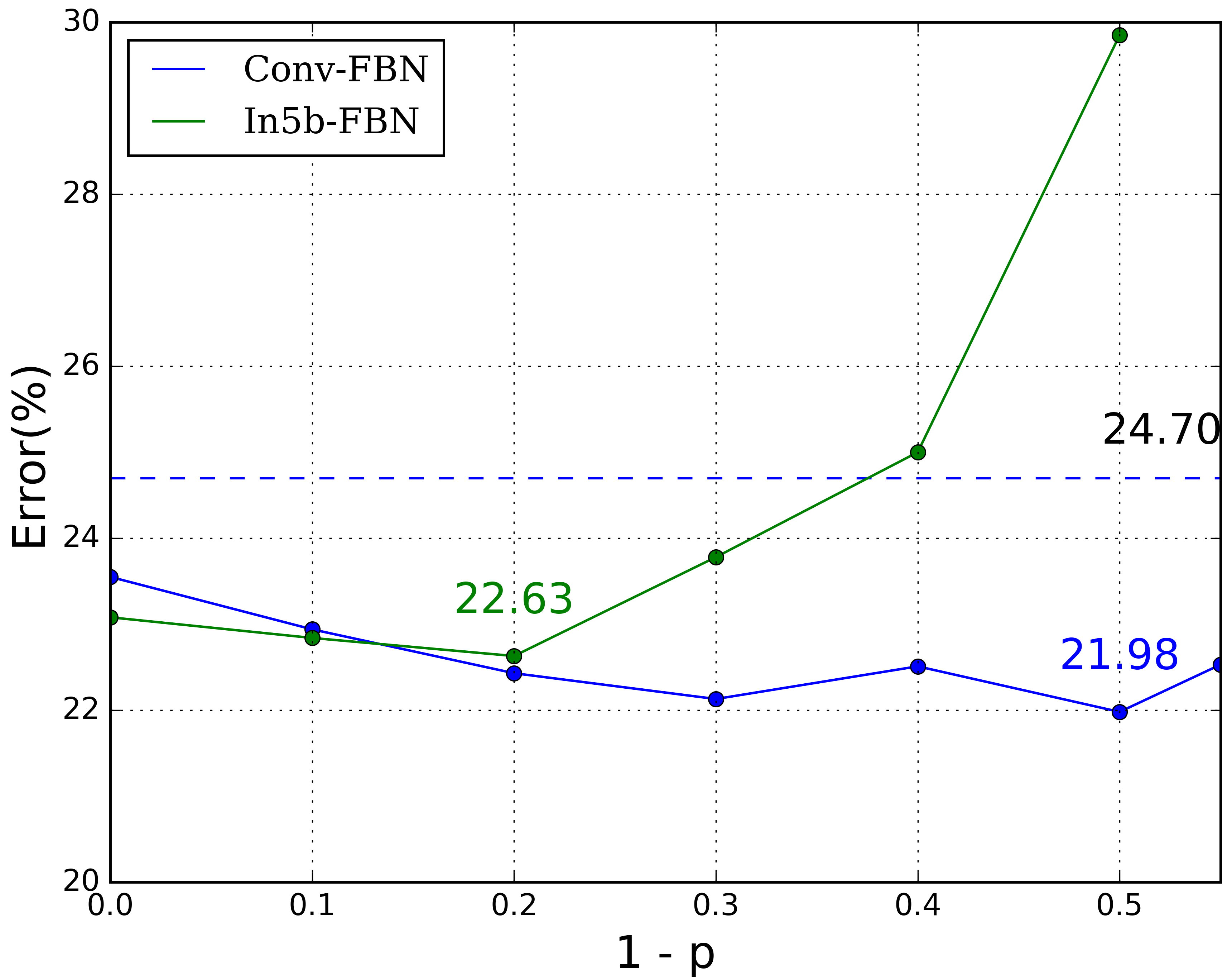}\label{fig:result-drop-a} }~~
\subfigure[Training curves of In5b-FBN] {\includegraphics[width=0.3\linewidth]{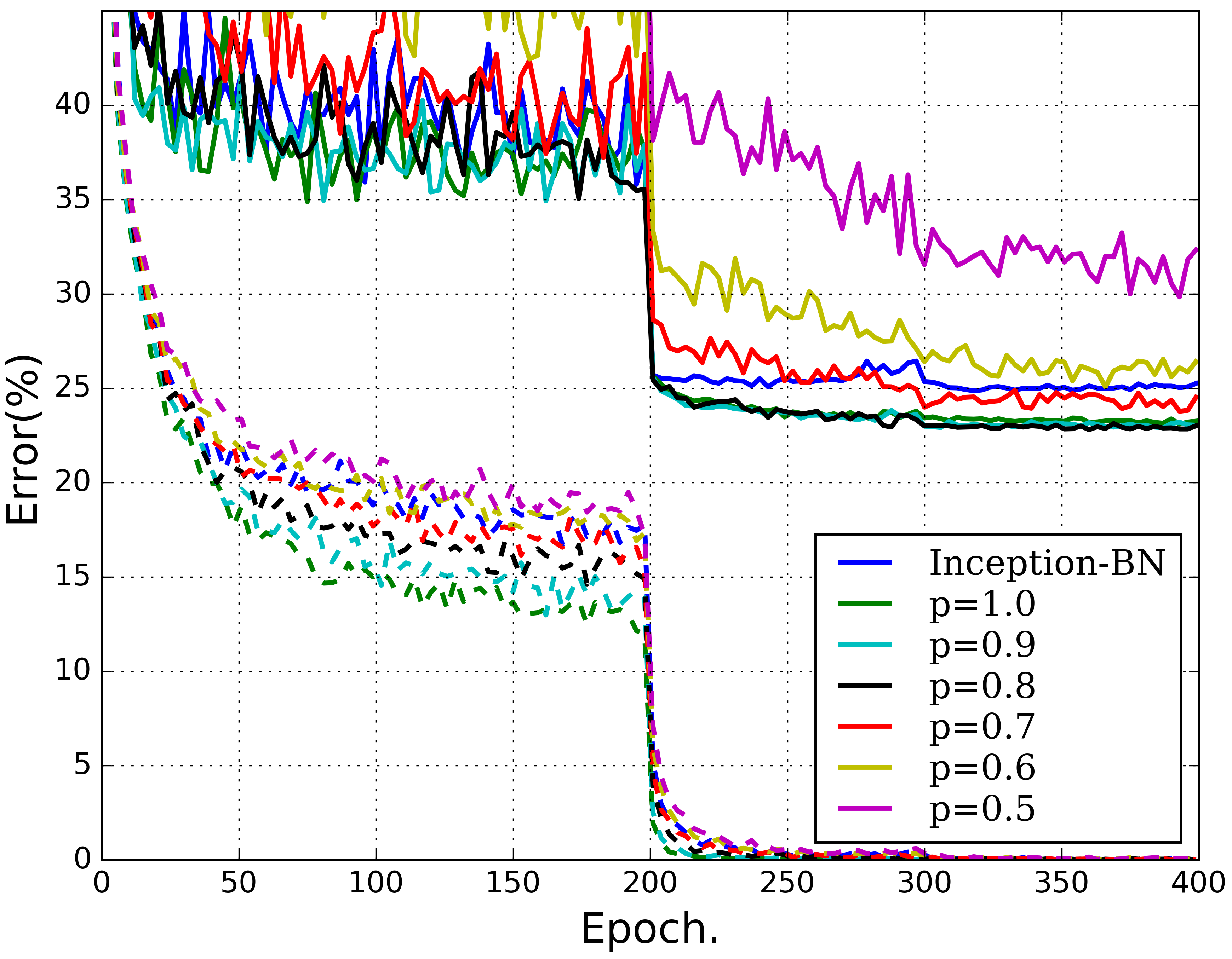}\label{fig:result-drop-b}}~~
\subfigure[Training curves of Conv-FBN] {\includegraphics[width=0.3\linewidth]{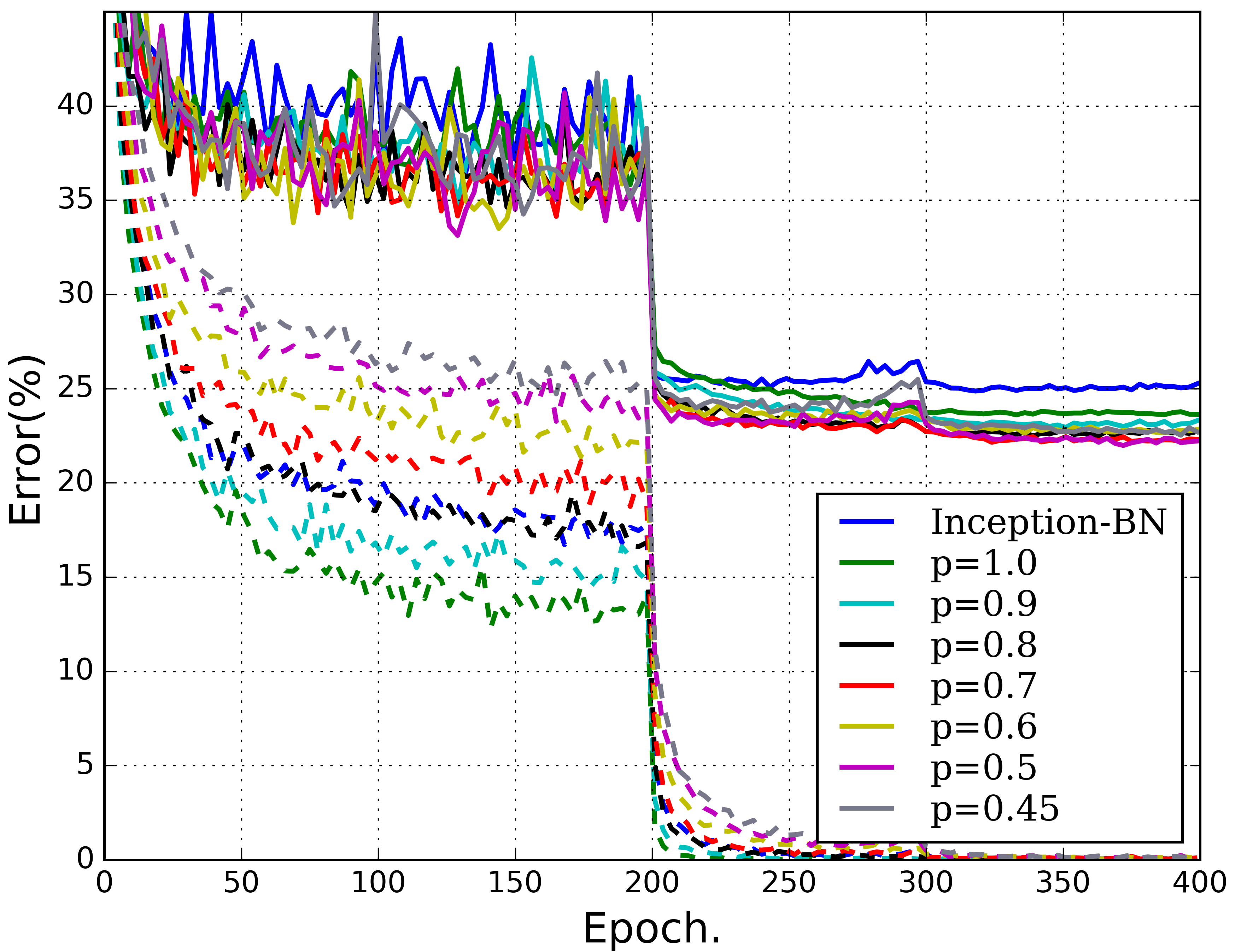}\label{fig:result-drop-c}}
}
\vspace{-1mm}
\caption{Training on CIFAR-100 of Conv-FBN networks with $k = 20$ and different $p$. (a) Test error (\%) on CIFAR-100 of In5b-FBN and Conv-FBN networks. (b)(c) Training curves of In5b-FBN and Conv-FBN networks. Dashed lines denote training error, and bold lines denote testing error. Note that we do not show the results of smaller DropFactor rates, since the performance drops significantly when $p$ is too small. Best viewed in color.}\label{fig:result-drop}
\vspace{-5.5mm}
\end{figure*}

\end{paragraph}
\begin{paragraph}{Speed Analysis of FB networks.} We show the runtime speed comparison of a small network (Inception-BN) and a relatively large network (ResNet of 1001 layers) with and without FB layers in Table~\ref{tbl:speed}. The test is performed on the Titan X GPU. Since the FB layers implemented on our own are not optimized by advanced implementation such as cuDNN~\cite{cudnn}, we also show the results of all methods without cuDNN for a fair comparison. For Inception-BN, the loss of speed is still tolerable. In addition, since we only insert a single FB block, it has little impact on the speed of large networks, \eg ResNet-1001. Lastly, cuDNN accelerates all methods a lot. We believe that the training speed of our FB layers will also benefit from a deliberate optimization.

\begin{table}[htbp]
\small
	\begin{center}
	\begin{tabular}{lcc}
		\hline	
								&  \multicolumn{2}{c}{Speed (samples/s)}	\\
		\hline
								& w/o cuDNN & cuDNN 	\\
		Inception-BN (24.70\%)			& 722.2 & 2231.1 \\
		Inception-BN-FBN	(21.98\%)	& 438.1 & 691.7\\
		\hline
		ResNet-1001	(20.50\%)			& 20.3 & 79.1\\
		ResNet-1001-FBN (19.67\%)		& 20.1 & 74.9\\
		\hline
	\end{tabular}
	\end{center}	
\vspace{-2mm}
	\caption{The training speeds of different methods on CIFAR-100. We use the Conv-FBN structure in the comparison.} \label{tbl:speed}
\vspace{-3mm}
\end{table}
\end{paragraph}

\end{subsection}

\begin{subsection}{Evaluation on Multiple Datasets}\label{exp:results}
In this section, we compare our FBN with other start-of-art methods on multiple datasets, including CIFAR-10, CIFAR-100, ImageNet and two fine-grained classification datasets. For the following experiments, we do not try exhaustive parameter search and use the Conv-FBN network with fixed factors $k=20$ and DropFactor rate $p=0.5$ as the default setting of FBNs, since this setting achieves the best performance according to the ablation experiments in Sec.~\ref{exp:ablation}. Our FB layers are implemented in MXNet~\cite{mxnet} and we follow some training policies in ``fb.resnet''\footnote{\label{fn:fbresnet}\url{https://github.com/facebook/fb.resnet.torch}}. % We will make the implementation public if the paper is accepted.

\begin{subsubsection}{Results on CIFAR-10 and CIFAR-100}
The CIFAR-10 and CIFAR-100~\cite{cifar} datasets contain 50,000 training images and 10,000 testing images of 10 and 100 classes, respectively. The resolution of each image is $32 \times 32$. We follow the moderate data augmentation in~\cite{he2016identity} for training: a random crop is taken from the image padded by 4 pixels or its horizontal flip. We use SGD for optimization with a weight decay of 0.0001 and momentum of 0.9. All models are trained with a minibatch size of 128 on two GPUs. For ResNet and its corresponding FBNs, we start training of a learning rate of 0.1 for total 200 epochs and divide it by 10 at 100 and 150 epochs. For Inception-BN based models, the learning rate is 0.2 at start and divided by 10 at 200 and 300 epochs for total 400 epochs.

We train three different networks: Inception-BN, ResNet-164 and ResNet-1001, and their corresponding FB networks. Note that we use the pre-activation version of ResNet in \cite{he2016identity} instead of the original ResNet~\cite{resnet}. Table~\ref{tbl:results} summarizes the results of our FBNs and other state-of-the-art algorithms. Our FBNs have consistent improvements over all three corresponding baselines. Specifically, our Inception-BN-FBN outperforms Inception-BN by 2.72\% on CIFAR-100 and 0.24\% on CIFAR-10, and ResNet-1001-FBN achieves the best result 19.67\% on CIFAR-100 among all the methods. A more intuitive comparison is in Fig.~\ref{fig:cifar100_bar}. Most remarkably, our method improves the performance with slightly additional cost of parameters. For example, compared to ResNet-1001 with 10.7M parameters, our ResNet-1001-FBN obtains better results with only 0.5M (5\%) additional parameters. This result is also better than the best Wide ResNet, which uses 36.5M parameters. Although Bilinear Pooling methods~\cite{lin2015bilinear,gao2015compact} were not utilized in general image classification tasks, we also re-implement them here using Inception-BN and ResNet-164 architectures. Their performance is inferior to our results.
\begin{table*}[htbp]
\small
	\begin{center}
	\begin{tabular}{lccc}
		\hline	
		Method 			& $\#$ params	& CIFAR-10 	& CIFAR-100	\\
		\hline	
		NIN~\cite{lin2013network}				& -			& 8.81		& 35.67\\
		DSN~\cite{lee2015deeply}				& -			& 8.22		& 34.57\\
		FitNet~\cite{romero2014fitnets}							& -			& 8.39		& 35.04\\
		Highway~\cite{highway}					& -			& 7.72		& 32.39\\
		ELU~\cite{clevert2015fast}				& -			& 6.55		& 24.28\\
		\hline
		Original ResNet-110~\cite{resnet}		& 1.7M		& 6.43		& 25.16\\
		Original ResNet-1202~\cite{resnet}	 	& 10.2M	& 7.93		& 27.82\\
		Stoc-depth-110~\cite{huang2016deep}		& 1.7M 				& 5.23		& 24.58\\
		Stoc-depth-1202~\cite{huang2016deep}		&10.2M 		& 4.91		& -	\\
		ResNet-164~\cite{he2016identity}			&1.7M		& 5.46		& 24.33 \\
		ResNet-1001~\cite{he2016identity}			&10.2M		& 4.62		& 22.71 \\
		FractalNet~\cite{larsson2016fractalnet}		& 22.9M		& 5.24      & 22.49 \\
		Wide ResNet (width$\times$8)~\cite{zagoruyko2016wide} &11.0M & 4.81  & 22.07\\
		Wide ResNet (width$\times$10)~\cite{zagoruyko2016wide} &36.5M & 4.17 & 20.50\\
		\hline
		Inception-BN-Bilinear~\cite{lin2015bilinear} & 13.1M 	& 5.82			& 25.72\\
		Inception-BN-TS~\cite{gao2015compact}       &  2.0M         & 5.75          & 24.63\\
		ResNet-164-Bilinear~\cite{lin2015bilinear}  & 8.3M 	    & 5.32  	& 23.85\\
		ResNet-164-TS~\cite{gao2015compact}         &  2.0M         & 5.58      & 23.48\\
		\hline
		Inception-BN 		& 1.7M 		&5.82			& 24.70\\
		ResNet-164 (ours)  		& 1.7M 		&5.30			& 23.64\\
		ResNet-1001	(ours)		& 10.2M		&\textbf{4.04}			& 20.50	\\
		Inception-BN-FBN    & 2.4M 		&5.58 			& 21.98\\
		ResNet-164-FBN		& 2.2M 		&5.00			& 22.50\\
		ResNet-1001-FBN		& 10.7M		&4.09			& \textbf{19.67}\\
		\hline
	\end{tabular}
	\end{center}
\vspace{-3mm}
	\caption{Top-1 error (\%) of different methods on CIFAR-10 and CIFAR-100 datasets using moderate data augmentation (flip/translation). The number of parameters is calculated on CIFAR-100.} \label{tbl:results}
\vspace{-5.5mm}
\end{table*}

\begin{figure}[htbp]
\center{
\includegraphics[width=0.65\linewidth]{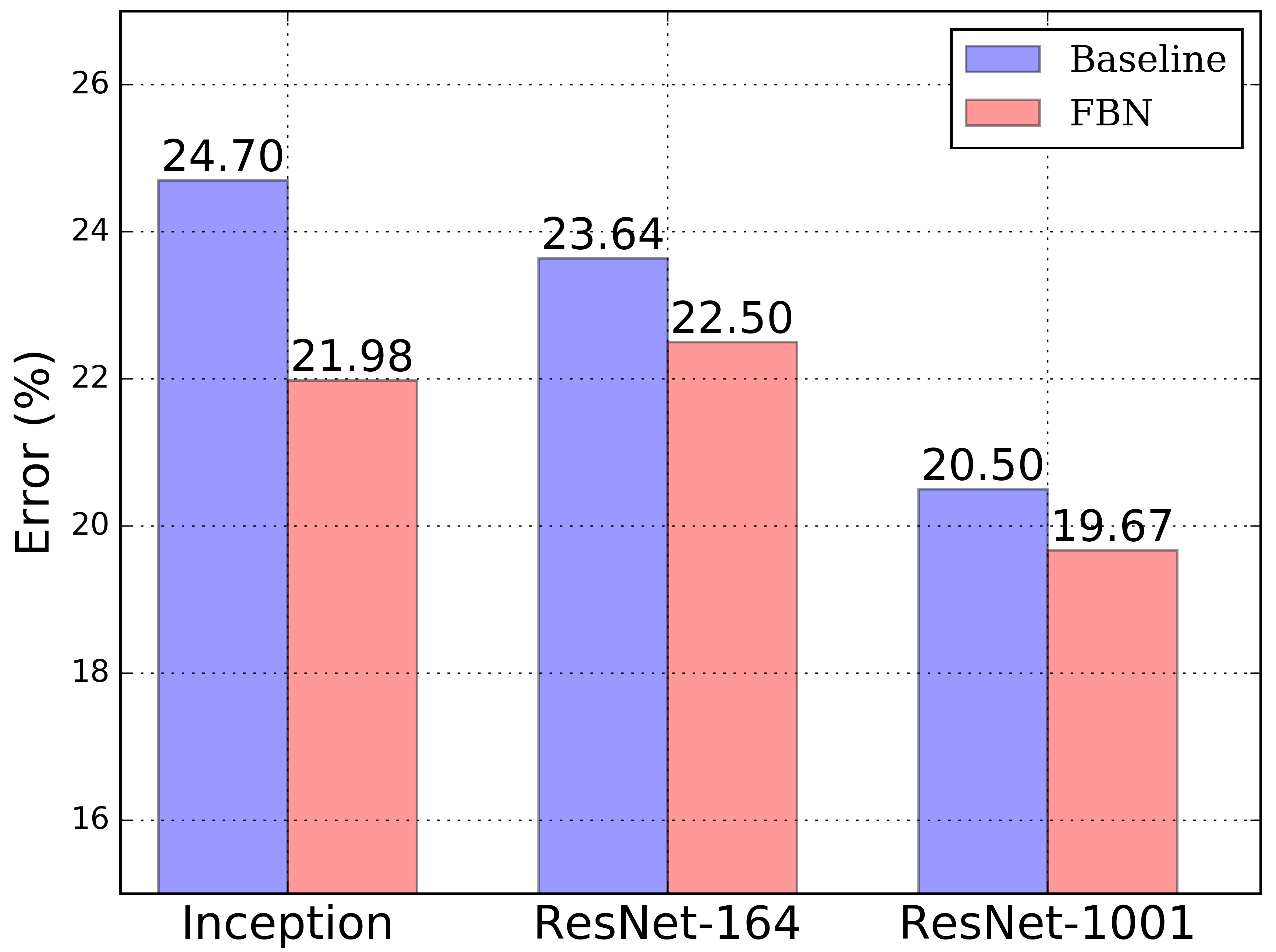}}
\vspace{-1.5mm}
\caption{Comparison of different baselines and their corresponding FBNs on CIFAR-100.\label{fig:cifar100_bar}}
\vspace{-6mm}
\end{figure}

\end{subsubsection}

\begin{subsubsection}{Results on ImageNet}
Although lots of works show their improvements on small datasets such as CIFAR-10 and CIFAR-100, few works prove their effectiveness in large scale datasets. Thus, in this section we evaluate our method on the ImageNet~\cite{imagenet} dataset, which is the golden test for image classification. The dataset contains 1.28M training images, 50K validation images and 100K testing images. We report the Top-1 and Top-5 errors of validation set in the single-model single center-crop setting. For the choice of FBN, we use the Conv-FBN structure in Sec.~\ref{exp:ablation} and the DropFactor rate is set as 0.5. In the training, we also follow some well-known strategies in ``fb.resnet''\footref{fn:fbresnet}, such as data augmentations and initialization method. The initial learning rate starts from 0.1 and is divided by 10 at 60, 75, 90 epochs for  120 epochs. 

\begin{table}[htbp]
\small
	\begin{center}
	\begin{tabular}{lcc}
		\hline	
		Method						& Top-1 (\%) 	& Top-5 (\%)	\\
		\hline
		Inception-BN~\cite{bn}		& 27.5	&	9.2\\
		Inception-BN-FBN				& 26.4	&   8.4 \\
		\hline
		ResNet-34~\cite{he2016identity} & 27.7	& 9.1\\
		ResNet-34-FBN				& 26.3	& 8.4\\
		\hline
		ResNet-50~\cite{he2016identity} & 24.7	& 7.4\\
		ResNet-50-FBN				& 24.0	& 7.1\\
		\hline
	\end{tabular}
	\end{center}
\vspace{-3mm}
	\caption{Comparisons of different methods by single center-crop error on the ImagNet validation set.} \label{tbl:imagenet}
\vspace{-6mm}
\end{table}

We adopt two modern network structures: Inception-BN~\cite{bn} and ResNet~\cite{he2016identity} in this experiment. %\footnote{We will report more results on the ResNet with deeper structures after the experiments are done.}. 
Table~\ref{tbl:imagenet} shows their results compared with FB variants. Relative to the original Inception-BN, our Inception-BN-FBN has a Top-1 error of 26.4\%, which is 1.1\% lower. ResNet-34-FBN and ResNet-50-FBN achieve 26.8\% and 24.7\% Top-1 error, and improve 1.4\% and 0.7\% over the baselines, respectively. The results demonstrate the effectiveness of our FB models on the large scale dataset.
\end{subsubsection}

\begin{subsubsection}{Results on Fine-grained Recognition Datasets}
Original Bilinear pooling methods~\cite{lin2015bilinear,gao2015compact} only show their results on fine-grained recognition applications, thus we apply our FB models in two fine-grained datasets CUB-200-2011~\cite{cub} and Describable Texture Dataset (DTD)~\cite{dtd} for comparisons. We use the same base network VGG-16 in this experiment. Table~\ref{tbl:fine} compares our method with bilinear pooling~\cite{lin2015bilinear} and two compact bilinear pooling~\cite{gao2015compact} methods (RM and TS). The results show that our FBN and the bilinear pooling methods all improves significantly over the VGG-16. We also re-implement bilinear pooling under the same training setting as our FBN. It should be more fair to compare its results (in the brackets) with our FBN. Note that our FBN also has much lower cost of memory and computation than bilinear pooling methods as described in Sec.~\ref{sec:relation}.
\begin{table}[t]
\small
	\begin{center}
	\begin{tabular}{lccccc}
		\hline	
		Dataset		& FC	& Bilinear~\cite{lin2015bilinear} & RM~\cite{gao2015compact} & TS~\cite{gao2015compact} & FBN	\\
		\hline
		CUB & 33.88 & \textbf{16.00} (17.79) & 16.14	& \textbf{16.00} & 17.09\\
		DTD & 39.89 & 32.50 (32.26) & 34.43 & 32.29 & \textbf{32.20}  \\
		\hline
	\end{tabular}
	\end{center}
\vspace{-3mm}
	\caption{Comparisons of different methods by classification error on CUB and DTD datasets. The number in the brackets are our re-implemented results.} \label{tbl:fine}
\vspace{-5.5mm}
\end{table}
\end{subsubsection}

\end{subsection}

% \begin{subsection}{Empirical Analysis of FBNs}\label{exp:visualization}
% \begin{paragraph}{Visualization of FBNs}
% XXX
% \end{paragraph}

% \end{subsection}

% \begin{subsection}{Fine-grained image classification}

% fine-grained ?

% \end{subsection}

\end{section}

\begin{section}{Conclusion and Future Work}
In this paper, we have presented the Factorized Bilinear (FB) model to incorporate pairwise interactions of features in neural networks. The method has low cost in both memory and computation, and can be easily trained in an end-to-end manner. To prevent overfitting, we have further proposed a specific regularization method \emph{DropFactor} by randomly dropping factors in FB layers. Our method achieves remarkable performance in several standard benchmarks, including CIFAR-10, CIFAR-100 and ImageNet. 

In the future work, we will go beyond the interactions inside features, and explore the generalization to model the correlations between samples in some more complicated tasks, such as face verification and re-identification.

\end{section}

\section*{Acknowledgement} This work was supported by the National Natural Science Foundation of China under Contract 61472011.

\pagebreak
{\small
\bibliographystyle{ieee}
\bibliography{egbib}

\begin{thebibliography}{10}\itemsep=-1pt

\bibitem{carreira2012semantic}
J.~Carreira, R.~Caseiro, J.~Batista, and C.~Sminchisescu.
\newblock Semantic segmentation with second-order pooling.
\newblock In {\em ECCV}, 2012.

\bibitem{chatfield2014return}
K.~Chatfield, K.~Simonyan, A.~Vedaldi, and A.~Zisserman.
\newblock Return of the devil in the details: Delving deep into convolutional
  nets.
\newblock {\em arXiv preprint arXiv:1405.3531}, 2014.

\bibitem{chen2015similarity}
D.~Chen, Z.~Yuan, G.~Hua, N.~Zheng, and J.~Wang.
\newblock Similarity learning on an explicit polynomial kernel feature map for
  person re-identification.
\newblock In {\em CVPR}, 2015.

\bibitem{mxnet}
T.~Chen, M.~Li, Y.~Li, M.~Lin, N.~Wang, M.~Wang, T.~Xiao, B.~Xu, C.~Zhang, and
  Z.~Zhang.
\newblock {MXNet}: A flexible and efficient machine learning library for
  heterogeneous distributed systems.
\newblock {\em NIPS Workshop on Machine Learning Systems}, 2016.

\bibitem{cudnn}
S.~Chetlur, C.~Woolley, P.~Vandermersch, J.~Cohen, J.~Tran, B.~Catanzaro, and
  E.~Shelhamer.
\newblock {cuDNN}: Efficient primitives for deep learning.
\newblock {\em arXiv preprint arXiv:1410.0759}, 2014.

\bibitem{chowdhury2016one}
A.~R. Chowdhury, T.-Y. Lin, S.~Maji, and E.~Learned-Miller.
\newblock One-to-many face recognition with bilinear cnns.
\newblock In {\em WACV}, 2016.

\bibitem{dtd}
M.~Cimpoi, S.~Maji, I.~Kokkinos, S.~Mohamed, and A.~Vedaldi.
\newblock Describing textures in the wild.
\newblock In {\em CVPR}, 2014.

\bibitem{clevert2015fast}
D.-A. Clevert, T.~Unterthiner, and S.~Hochreiter.
\newblock Fast and accurate deep network learning by exponential linear units
  ({ELUs}).
\newblock {\em arXiv preprint arXiv:1511.07289}, 2015.

\bibitem{gao2015compact}
Y.~Gao, O.~Beijbom, N.~Zhang, and T.~Darrell.
\newblock Compact bilinear pooling.
\newblock In {\em CVPR}, 2016.

\bibitem{rcnn}
R.~Girshick, J.~Donahue, T.~Darrell, and J.~Malik.
\newblock Rich feature hierarchies for accurate object detection and semantic
  segmentation.
\newblock In {\em CVPR}, 2014.

\bibitem{he2015delving}
K.~He, X.~Zhang, S.~Ren, and J.~Sun.
\newblock Delving deep into rectifiers: Surpassing human-level performance on
  imagenet classification.
\newblock In {\em CVPR}, 2015.

\bibitem{resnet}
K.~He, X.~Zhang, S.~Ren, and J.~Sun.
\newblock Deep residual learning for image recognition.
\newblock {\em CVPR}, 2016.

\bibitem{he2016identity}
K.~He, X.~Zhang, S.~Ren, and J.~Sun.
\newblock Identity mappings in deep residual networks.
\newblock {\em ECCV}, 2016.

\bibitem{hornik1989multilayer}
K.~Hornik, M.~Stinchcombe, and H.~White.
\newblock Multilayer feedforward networks are universal approximators.
\newblock {\em Neural Networks}, 2(5):359--366, 1989.

\bibitem{huang2016deep}
G.~Huang, Y.~Sun, Z.~Liu, D.~Sedra, and K.~Weinberger.
\newblock Deep networks with stochastic depth.
\newblock {\em ECCV}, 2016.

\bibitem{bn}
S.~Ioffe and C.~Szegedy.
\newblock Batch normalization: Accelerating deep network training by reducing
  internal covariate shift.
\newblock In {\em ICML}, 2015.

\bibitem{jia2014learning}
Y.~Jia.
\newblock {\em Learning Semantic Image Representations at a Large Scale}.
\newblock PhD thesis, EECS Department, University of California, Berkeley,
  2014.

\bibitem{kar2012random}
P.~Kar and H.~Karnick.
\newblock Random feature maps for dot product kernels.
\newblock In {\em AISTATS}, 2012.

\bibitem{kontschieder2015deep}
P.~Kontschieder, M.~Fiterau, A.~Criminisi, and S.~Rota~Bulo.
\newblock Deep neural decision forests.
\newblock In {\em CVPR}, 2015.

\bibitem{cifar}
A.~Krizhevsky.
\newblock Learning multiple layers of features from tiny images.
\newblock {\em Tech Report}, 2009.

\bibitem{alexnet}
A.~Krizhevsky, I.~Sutskever, and G.~E. Hinton.
\newblock {ImageNet} classification with deep convolutional neural networks.
\newblock In {\em NIPS}, 2012.

\bibitem{larsson2016fractalnet}
G.~Larsson, M.~Maire, and G.~Shakhnarovich.
\newblock {FractalNet}: Ultra-deep neural networks without residuals.
\newblock {\em arXiv preprint arXiv:1605.07648}, 2016.

\bibitem{lecun1989backpropagation}
Y.~LeCun, B.~Boser, J.~S. Denker, D.~Henderson, R.~E. Howard, W.~Hubbard, and
  L.~D. Jackel.
\newblock Backpropagation applied to handwritten zip code recognition.
\newblock {\em Neural Computation}, 1(4):541--551, 1989.

\bibitem{lee2015deeply}
C.-Y. Lee, S.~Xie, P.~Gallagher, Z.~Zhang, and Z.~Tu.
\newblock Deeply-supervised nets.
\newblock In {\em AISTATS}, 2015.

\bibitem{lin2013network}
M.~Lin, Q.~Chen, and S.~Yan.
\newblock Network in network.
\newblock In {\em ICLR}, 2014.

\bibitem{lin2015bilinear}
T.-Y. Lin, A.~RoyChowdhury, and S.~Maji.
\newblock Bilinear {CNN} models for fine-grained visual recognition.
\newblock In {\em CVPR}, 2015.

\bibitem{fcn}
J.~Long, E.~Shelhamer, and T.~Darrell.
\newblock Fully convolutional networks for semantic segmentation.
\newblock In {\em CVPR}, 2015.

\bibitem{relu}
V.~Nair and G.~E. Hinton.
\newblock Rectified linear units improve restricted {Boltzmann} machines.
\newblock In {\em ICML}, 2010.

\bibitem{pham2013fast}
N.~Pham and R.~Pagh.
\newblock Fast and scalable polynomial kernels via explicit feature maps.
\newblock In {\em SIGKDD}, 2013.

\bibitem{ren2015vectorization}
J.~S. Ren and L.~Xu.
\newblock On vectorization of deep convolutional neural networks for vision
  tasks.
\newblock In {\em AAAI}, 2015.

\bibitem{faster-rcnn}
S.~Ren, K.~He, R.~Girshick, and J.~Sun.
\newblock Faster {R-CNN}: Towards real-time object detection with region
  proposal networks.
\newblock In {\em NIPS}, 2015.

\bibitem{rendle2010factorization}
S.~Rendle.
\newblock Factorization machines.
\newblock In {\em ICDM}, 2010.

\bibitem{romero2014fitnets}
A.~Romero, N.~Ballas, S.~E. Kahou, A.~Chassang, C.~Gatta, and Y.~Bengio.
\newblock Fitnets: Hints for thin deep nets.
\newblock In {\em ICLR}, 2015.

\bibitem{imagenet}
O.~Russakovsky, J.~Deng, H.~Su, J.~Krause, S.~Satheesh, S.~Ma, Z.~Huang,
  A.~Karpathy, A.~Khosla, M.~Bernstein, et~al.
\newblock {ImageNet} large scale visual recognition challenge.
\newblock {\em International Journal of Computer Vision}, 115(3):211--252,
  2015.

\bibitem{shawe2004kernel}
J.~Shawe-Taylor and N.~Cristianini.
\newblock {\em Kernel methods for pattern analysis}.
\newblock Cambridge University Press, 2004.

\bibitem{dropout}
N.~Srivastava, G.~E. Hinton, A.~Krizhevsky, I.~Sutskever, and R.~Salakhutdinov.
\newblock Dropout: a simple way to prevent neural networks from overfitting.
\newblock {\em Journal of Machine Learning Research}, 15(1):1929--1958, 2014.

\bibitem{highway}
R.~K. Srivastava, K.~Greff, and J.~Schmidhuber.
\newblock Training very deep networks.
\newblock In {\em NIPS}, 2015.

\bibitem{googlenet}
C.~Szegedy, W.~Liu, Y.~Jia, P.~Sermanet, S.~Reed, D.~Anguelov, D.~Erhan,
  V.~Vanhoucke, and A.~Rabinovich.
\newblock Going deeper with convolutions.
\newblock In {\em CVPR}, 2015.

\bibitem{cub}
C.~Wah, S.~Branson, P.~Welinder, P.~Perona, and S.~Belongie.
\newblock The {C}altech-{UCSD} {B}irds-200-2011 dataset.
\newblock 2011.

\bibitem{zagoruyko2016wide}
S.~Zagoruyko and N.~Komodakis.
\newblock Wide residual networks.
\newblock {\em arXiv preprint arXiv:1605.07146}, 2016.

\bibitem{zeiler2014visualizing}
M.~D. Zeiler and R.~Fergus.
\newblock Visualizing and understanding convolutional networks.
\newblock In {\em ECCV}, 2014.

\end{thebibliography}
}

\pagebreak

%\begin{section}{Supplementary}

\begin{section}{Supplementary: More Exploration Experiments of Factorized Bilinear Models}

We study the effect of different kernel sizes in Factorized Bilinear (FB) Models. We also present comparisons with Dropout and our DropFactor in FB models.

\begin{subsection}{Effect of Kernel Size}
Tab.~\ref{tbl:large} shows the results of different kernel sizes (1x1 and 3x3) for FB layers. We conduct experiments on CIFAR-100 dataset~\cite{cifar} with two FB networks In5b-FBN and Conv-FBN as described in the paper. We insert a 1x1 FB layer and 3x3 FB layer, respectively, for both two FBNs.  The results of 3x3 kernel size are still better than the baseline. This demonstrates that our FB models can generalize to model interactions with larger kernel size. However, it also leads to more severe over-fitting than 1x1 at least on CIFAR-100 and has 9 times parameters than an 1x1 FB layer. Thus, incorporating 1x1 FB layer can achieve more efficient and effective performance.
% \begin{table}[htbp]
% 	\begin{center}
% 	\begin{tabular}{c|c|cc|cc}
% 		\hline	
% 		\multirow{2}{*}{Method} & \multirow{2}{*}{Inception-BN}	 & \multicolumn{2}{c}{In5b-FBN} & \multicolumn{2}{c}{Conv-FBN}	\\
% 		\cline{3-6}
% 		 & & 1x1 & 3x3 & 1x1 & 3x3  \\
% 		\hline
% 		Error & 24.70 & 22.63 & 23.87 & 21.98 & 23.08 \\
% 		\hline
% 	\end{tabular}
% 	\end{center}
% 	\caption{Different kernel size of FB layers on CIFAR-100.} \label{tbl:large}
% \end{table}

\begin{table}[htbp]
	\begin{center}
	\begin{tabular}{ccc}
		\hline	
		Method & Kernel Size & Error		\\
		\hline
		Inception-BN & - & 24.70\\
		\hline
		\multirow{2}{*}{In5b-FBN} & 1x1 & 22.63\\
		& 3x3 & 23.87 \\
		\hline
		\multirow{2}{*}{Conv-FBN} & 1x1 & 21.98\\
		& 3x3 & 23.08 \\
		\hline
	\end{tabular}
	\end{center}
	\caption{Results of different kernel sizes in the FB layers on the CIFAR-100 dataset.} \label{tbl:large}
\end{table}

\end{subsection}

\begin{subsection}{Comparisons with Dropout and DropFactor} 
\begin{table}[htbp]
	\begin{center}
	\begin{tabular}{clc}
		\hline	
		Network & {Method}	 & Error \\
		\hline
		\multirow{4}{*}{Inception-BN} & Baseline & 24.70\\
		& FBN & 23.55 \\
		&FBN + Dropout & 23.19 \\
		&FBN + DropFactor & \textbf{21.98} \\
		&FBN + Dropout + DropFactor & 22.71 \\
		\hline
		\multirow{4}{*}{ResNet-164} & Baseline & 23.64\\
		& FBN & 23.39 \\
		&FBN + Dropout & 22.97 \\
		&FBN + DropFactor & \textbf{22.50} \\
		&FBN + Dropout + DropFactor & 22.60 \\
		\hline
	\end{tabular}
	\end{center}
	\caption{Results of Dropout and DropFactor on the CIFAR-100 dataset.} \label{tbl:drop}
\end{table}
Our DropFactor scheme shares similar idea with Dropout~\cite{dropout}, which is also a simple yet effective regularization to prevent over-fitting. We evaluate the performance of Dropout with our specific designed DropFactor for Factorized Bilinear models. Tab.~\ref{tbl:drop} illustrates the results of two methods on the CIFAR-100 dataset. We adopt the Inception-BN and ResNet-164 networks as the base networks in this experiments. The FBN models are constructed by inserting the FB layers in the base networks. As shown in the table, Dropout and DropFactor both improve the performance individually over the original FBN model. DropFactor achieves even better results and combining them does not get further improvement. This demonstrates the effectiveness of DropFactor scheme to reduce the over-fitting of FB models. 
% \begin{table}[htbp]
% 	\begin{center}
% 	\begin{tabular}{ccccc}
% 		\hline	
% 		\multirow{2}{*}{Method}	 & \multirow{2}{*}{FBN} & FBN + & FBN + & FBN + dropout +	\\
% 		 & & dropout & DropFactor & DropFactor \\
% 		\hline
% 		Error &  23.55 & 23.19 & \textbf{21.98}  & 22.71 \\
% 		\hline
% 	\end{tabular} 
% 	\end{center}
% 	\caption{Results of Dropout and DropFactor on  CIFAR-100.} \label{tbl:drop}
% \end{table}

\end{subsection}

\end{section}

%\end{section}

\end{document}